\documentclass[10pt,journal,cspaper,compsoc]{IEEEtran}

\usepackage{graphicx}
\usepackage{amsmath,amsfonts,graphicx,url}
\usepackage{array}
\usepackage{color}
\usepackage{algorithm,algorithmic}

\newcommand{\X}{\mathbf{X}}
\newcommand{\A}{\mathbf{A}}
\newcommand{\B}{\mathbf{B}}
\newcommand{\x}{\mathbf{x}}
\newcommand{\Y}{\mathbf{Y}}
\newcommand{\y}{\mathbf{y}}

\newcommand{\w}{\mathbf{w}}
\newcommand{\W}{\mathbf{W}}
\newcommand{\Z}{\mathbf{Z}}

\newcommand{\I}{\mathbf{I}}
\newcommand{\N}{\mathcal{N}}
\newcommand{\G}{\mathcal{G}}

\newcommand{\z}{\mathbf{z}}

\newcommand{\U}{\mathbf{U}}
\newcommand{\V}{\mathbf{V}}
\newcommand{\uu}{\mathbf{u}}
\newcommand{\vv}{\mathbf{v}}
\newcommand{\muu}{\boldsymbol{\mu}_u}
\newcommand{\muv}{\boldsymbol{\mu}_v}
\newcommand{\Alp}{\boldsymbol{\alpha}}
\newcommand{\Sig}{\boldsymbol{\Sigma}}

\newcommand{\Sigmab}{\boldsymbol{\Sigma}}
\newcommand{\eps}{\boldsymbol{\epsilon}}

\newcommand{\Ta}{\boldsymbol{\tau}}
\newcommand{\m}{\mathbf{m}}

\begin{document}

\title{Group Factor Analysis}

\author{Arto~Klami,
		Seppo~Virtanen,
        Eemeli~Lepp\"{a}aho,
        and~Samuel~Kaski
\IEEEcompsocitemizethanks{
\IEEEcompsocthanksitem A. Klami and S. Kaski are with Helsinki Institute for Information Technology HIIT, Department of Computer Science,  University of Helsinki.\protect\\
E-mail: arto.klami@hiit.fi, samuel.kaski@hiit.fi
\IEEEcompsocthanksitem S. Virtanen, E. Lepp\"{a}aho and S. Kaski are with Helsinki Institute for Information Technology HIIT, Department of Information and Computer Science, Aalto University.\protect\\
E-mail: seppo.j.virtanen@aalto.fi,eemeli.leppaaho@aalto.fi}
\thanks{}}

\date{Received: date / Accepted: date}

\maketitle

\begin{abstract}
  Factor analysis provides linear factors that describe relationships
  between individual variables of a data set. We extend this classical
  formulation into linear factors that describe relationships between
  groups of variables, where each group represents either a set of
  related variables or a data set. The model also naturally extends
  canonical correlation analysis to more than two sets, in a way that
  is more flexible than previous extensions. Our solution is formulated
  as variational inference of a latent variable model with structural
  sparsity, and it consists of two hierarchical levels: The higher
  level models the relationships between the groups, whereas the lower
  models the observed variables given the higher level. We show that
  the resulting solution solves the group factor analysis problem accurately,
  outperforming alternative factor analysis based solutions as well
  as more straightforward implementations of group factor analysis.
  The method is demonstrated on two life science data sets, one on
  brain activation and the other on systems biology, illustrating its
  applicability to the analysis of different types of high-dimensional
  data sources.

\end{abstract}

\section{Introduction}

Factor analysis (FA) is one of the cornerstones of data analysis, the
tool of choice for capturing and understanding linear relationships
between variables \cite{Thurstone31}. It provides a set of $K$
factors, each explaining dependencies between
some of the features in a vectorial data sample $\y_i \in \mathbb{R}^D$ based on the
model
\[
\y_i = \sum_{k=1}^K z_{i,k} \w_k + \eps_i,
\]
where $z_{i,k}$ is the value of the $k$th unobserved factor, $\w_k \in
\mathbb{R}^D$ contains its loadings, and $\eps_i$ is Gaussian
noise. To correctly capture the relationships, we need to assume a
diagonal noise covariance with free variance for each of the
variables. If the noise model was more flexible, having non-diagonal
covariance, it would allow describing some of the relationships as
noise.  On the other hand, forcing the variances to be equal would
imply that heteroscedastic noise would need to be explained as
factors, reducing the model to probabilistic PCA \cite{Bishop99nips}.

Building on our preliminary conference paper \cite{Virtanen12aistats},
we generalize factor analysis to a novel problem formulation of
\emph{group factor analysis} (GFA), where the task is to explain
relationships between groups of variables. We retain the
linear-Gaussian family of FA, but modify the model so that each factor now
describes dependencies between some of the feature groups instead of
individual variables. Again the
choice of residual noise is crucial: it needs to be flexible
enough to model everything that is not a true relationship between
two variable groups, but restricted enough so that all actual
relationships will be modeled as individual factors. For FA these
requirements were easily satisfied by assuming independent variance
for each dimension. For GFA more elaborate constructions are needed,
but the same basic idea applies.

From another perspective, GFA extends multi-battery factor analysis
(MBFA), introduced by McDonald \cite{McDonald70} and Browne
\cite{Browne80} as a generalization of inter-battery factor analysis
(IBFA) \cite{Tucker58,Browne79} to more than two variable groups. MBFA is a
factor analysis model for multiple co-occurring data sets, or,
equivalently, for a vectorial data sample whose variables have been split
into groups.  It includes a set of factors that model the
relationships between all variables, as well as separate sets of
factors explaining away the noise in each of the variable groups.  These
group-specific factor sets are sufficiently flexible for modeling
all variation within each group. However, each of the remaining
factors is assumed to describe relationships between \emph{all} of the
groups, which is not sufficient for providing interpretable factors
that reveal the relationships between the data sets as will be
explained below. Nevertheless, the MBFA models are useful tools for
multi-source data analysis, illustrated by the fact that the problem
has been re-discovered in machine learning literature several times;
see Section~\ref{sec:related} for more details.

To solve the GFA problem, we need to have also factors that describe
relationships between subsets of the groups. This makes the solutions
to the problem both more flexible and more interpretable than MBFA. For example,
a strong factor tying two groups while being independent of the other
groups can then be explicitly modeled as such. The MBFA-based models
would, falsely, reveal such a factor as one that is shared by all
groups. Alternatively, they would need to, again incorrectly, split them
into multiple group-specific ones.

In recent years, the need for the GFA solution 
has been identified by several authors, under different
terminology. Jia et al.  learned sparse matrix
factorization by convex optimization \cite{Jia10}, and Van
Deun et al. used group-lasso penalty to constrain the factors of a
simultaneous component analysis (SCA) model \cite{VanDeun11}. Various
Bayesian techniques have also been proposed for learning shared and
individual subspaces of multiple data sources
\cite{Virtanen12aistats,Gupta11,Gupta12,Damianou12}.

In this work we lay the foundation for future development of GFA
solutions, by properly defining the problem setup and terminology.  We
also present a general solution outline and show that the solutions
mentioned above are all instances of the same basic approach; they all
learn structured sparse FA models with varying techniques for
obtaining group-wise sparsity for the factor loadings. We then
propose a novel GFA solution that does not make a strong simplifying assumption shared by all
the previous approaches. They all assume that we can independently infer,
for each factor-group pair, whether that factor describes
variation related to that group, whereas our
solution explicitly models also these
associations with an additional linear model. In brief, our model
hence consists of two linear hierarchical levels. The first models the
relationships between the groups, and the latter models the observed
data given the output of the higher level. Alternatively, it can be
viewed as a direct generalization of \cite{Virtanen12aistats} with
a more advanced structured sparsity prior making it possible to reduce the
degrees of freedom in the model when needed.

Before delving into the details on how we solve the GFA problem, we
introduce some general application scenarios. The model is useful 
for analyzing multi-view setups where we have several data
sets with co-occurring samples. The variables can be grouped according
to the data sets: all variables in one set belong to one group
etc. Then GFA explains relationships between data sources,
and for two data sets it equals the problem of canonical correlation
analysis (CCA; see \cite{Klami13jmlr} for a recent overview from a
probabilistic perspective).  Alternatively, each group could contain a collection of
variables chosen to represent a multi-dimensional concept, such as
cognitive abilities of a subject, which cannot be summarized with a
single feature. Then GFA could be used for associating cognitive
abilities with other multi-dimensional concepts. The groups can also
represent a meaningful partitioning of larger data sets; we present two
practical examples of this kind of a setup. In one example we split a
high-dimensional feature vector over the human genome into subsets
according to functional pathways to describe drug responses, and in
the other example we split magnetic resonance images of the human brain
into local regions to study relationships between brain areas.

\section{Group factor analysis}

\subsection{Problem formulation}

The group factor analysis problem is as follows: Assume a collection of observations $\y_i\in \mathbb{R}^D$ for $i=1,\dots,N$ collected in a data matrix
$\Y \in \mathbb R^{N \times D}$, and a disjoint partition of the $D$
variables into $M$ groups $\{ G_m\}$. The GFA task is to find a set of $K$ factors that
describe $\Y$ so that relationships between the groups can be
separated from relationships within the groups. For notational
simplicity, assume that the first $D_1$ variables correspond to the
first group $G_1$, the following $D_2$ variables to $G_2$, and so
on. Then we can write $\Y = [\X^{(1)},...,\X^{(M)}]$, where $\X^{(m)}$ is a subset
of the data corresponding to $G_m$. We use 
$\x^{(m)}_i$ to denote the $i$th sample (row) of $\X^{(m)}$.
Throughout this paper we use the superscript $^{(m)}$ to denote variables related to the $m$th group or data set.

\begin{figure*}[!t]
  \centering
  \def\svgwidth{450pt}
  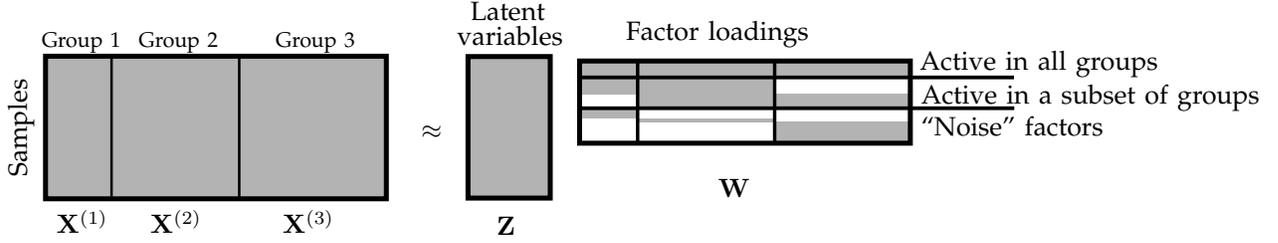
  \caption{Illustration of the group factor setup for three groups. The model learns a linear factorization of a data matrix $\Y=[\X^{(1)},\X^{(2)},\X^{(3)}]$ whose features have been split into (here) three groups, so that the factor loadings $\W$ are group-wise sparse. The model then automatically learns which factors describe dependencies between either all of the groups or a subset of them, and which describe structured noise specific to each group. The sparsity structure of $\W$ is here represented by coloring; the white areas correspond to zeros whereas the gray areas are non-zero.}
  \label{fig:factorization}
\end{figure*}

\subsection{General solution}
\label{sec:general}

A general solution to the GFA problem can be formulated
as a joint factor model for the observed data sets. 
The model for the $m$th
group of the $i$th sample is
\begin{align}
\x^{(m)}_{i} \sim \N( \W^{(m)\top} \z_{i}, \tau_m^{-1} \I ), \label{eq:model}
\end{align}
where $\W^{(m)^\top} = [\w_1^{(m)},\dots,\w_{K}^{(m)}]$
, $\z_i \in \mathbb{R}^{K}$, and $\tau_m$
is noise precision. 
Equivalently, we can directly write $\y_i = \W^\top \z_i + \eps_i$,
where $\eps_i$ is Gaussian noise with diagonal covariance but separate
variance for each group, by
denoting $\W = [\W^{(1)},...,\W^{(M)}]$.  

To make the factors
interpretable in the GFA-sense, that is, to describe relationships
between the groups, we need to make $\W$ sparse so that it satisfies
the following properties (for a visual illustration see Figure \ref{fig:factorization}):
\begin{enumerate}
\item Some factors are private to each group, so that $\w^{(m)}_{k} \neq 0$
only for one $m$. These factors explain away the variation independent
of all the other groups, and play the role of residual noise in regular FA.
\item The rest of the factors describe relationships between some
arbitrary subset of the groups; they are non-zero for those
groups and zero for the others.
\end{enumerate}

A trivial solution would explicitly split the factors into separate
sets so that there would be one set of factors for each possible
subset of the groups (including the singletons and the set of all
groups).  This can be done for small $M$; for example Klami and Kaski
proposed such a model for $M=2$ \cite{Klami08neuroc} and Gupta et
al. formulated the model for general $M$ but ran experiments only with
$M=3$ \cite{Gupta11}. Due to the exponential number of subsets, these
approaches cannot generalize to large $M$.

A better approach is to associate the projection matrix $\W$ with a
structural sparsity prior that encourages solutions that satisfy the
necessary properties. This strategy was first presented for $M=2$
by Virtanen et al. \cite{Virtanen11}, and extended for general $M$ independently
by several authors \cite{Virtanen12aistats,Gupta12,Damianou12}.
Despite technical differences in how the structural sparsity is
obtained, all of these approaches can be seen as special instances
of our GFA solution principle. Also the non-Bayesian models that
can be used to solve the GFA problem follow the same principle
\cite{Jia10,VanDeun11}.

\section{Model}

We propose a novel GFA solution that is another instantiation of the
general approach described above. The technical novelty is in a more
advanced structural sparsity prior which takes into account possible
dependencies between the groups, instead of assuming the group-factor
activations to be \emph{a priori} independent as in the earlier
solutions. The model can also be interpreted as a two-level
model that uses one level to model association strengths between
individual groups and the other level to model the observations given
the association strength. This interpretation clarifies the conceptual
novelty, explicating how the new structural sparsity prior has an intuitive
interpretation.

The generative model is the one given in \eqref{eq:model} coupled
with suitable priors. For $\z_i$ we use the unit Gaussian
prior $\z_i \sim \N(\mathbf{0},\I)$, and for the noise precisions $\tau_m$ we
employ a gamma prior with both shape and rate parameters set to $10^{-14}$; the model is
fairly insensitive to these hyperparameters. To find a GFA solution
these standard choices need to be complemented with structured
sparse priors for $\W$, described next.

\subsection{Sparsity prior}

We denote by $\alpha_{m,k}$ the inverse strength of association between the $m$th
group and the $k$th factor, and directly interpret it
 as the precision parameter of the prior distribution
for $\w_k^{(m)}$, the projection mapping the $k$th factor to
the observed variables in the $m$th group. That is, we assume
the prior
\[
p(\W|\Alp) = \prod_{m=1}^M \prod_{k=1}^{K} \prod_{d=1}^{D_m} \N(\w^{(m)}_{k,d} | 0,\alpha_{m,k}^{-1}).
\]
The same prior  was used in our preliminary work
\cite{Virtanen12aistats}, where we drew $\alpha_{m,k}$ independently
from a flat gamma prior to implement group-wise extension to
automatic relevance determination (ARD).

\begin{figure}[!t]
  \centering
  \def\svgwidth{230pt}
  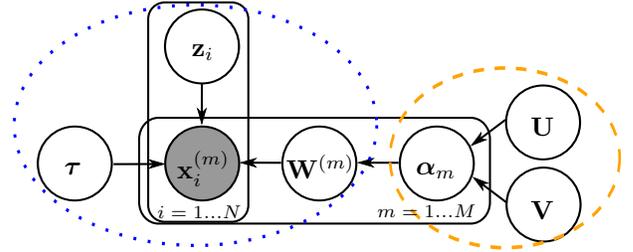
  \caption{Plate diagram of group factor analysis. The observation model, used also
by earlier GFA solutions, is
highlighted by the blue dotted region, whereas the novel low-rank model for the group-factor
associations is indicated by the orange dashed region.}
  \label{fig:gfaplate}
\end{figure}

Here we replace the independent draws with a linear model for $\Alp$
to explicitly model the association strengths between group-factor
pairs.  Since the entries correspond to precisions for the second
level projections, we model them in the log-space as
\begin{equation}
\log \Alp = \U \V^\top + \muu \mathbf{1}^\top + \mathbf{1}\muv^\top, \label{eq:lrmodel}
\end{equation}
where $\U\in\mathbb{R}^{M\times R}$ and $\V \in \mathbb{R}^{K\times R}$. The
vectors $\muu \in \mathbb{R}^M$ and $\muv \in \mathbb{R}^K$ model the mean
profiles.
Here $R$ is the rank of the linear model, and typically
$R \ll \min(M,K)$ so that we get a low-rank decomposition for
the association strengths, obtained by element-wise exponentiation
$\Alp = \exp (\U \V^\top + \muu \mathbf{1}^\top + \mathbf{1}\muv^\top )$.
Finally, we place an element-wise normal prior for the matrices $\U$ and $\V$
with zero mean and precision set to a fixed constant
$\lambda=0.1$; extensions to further hierarchical priors would also be
tractable if needed.
The resulting GFA model is visualized as a plate diagram in Figure \ref{fig:gfaplate},
highlighting the two levels of the model.

The motivation for modeling the $\alpha_{m,k}$ instead of assuming them independent
comes from the original modeling task
of GFA. The goal is to understand the relationships between the groups,
and hence we should explicitly model them. The earlier models
with independent priors assume that the groups are independent, which
is unlikely to hold in practical applications. Our model, in turn,
directly represents correlations between the group activation profiles.

An alternative formulation for correlated groups would
directly draw $\log \Alp_{m}$ from a multivariate distribution,
such as multivariate normal \cite{Aitchison82}. However, specifying the correlations
for such a model would require $M(M-1)/2$ parameters, making the
approach infeasible for large $M$. Since modeling the correlations
is expected to be the most useful for large number of groups, it is
clearly beneficial to use the low-rank model that requires only
$(M+K) \times (R+1)$ parameters.

\subsection{Interpretation}

As mentioned above, the model can be interpreted in two alternative
ways. The straightforward interpretation is that of a factor analysis
model for the $D$ observed variables, with a structural sparsity prior
for making the projections implement the GFA properties. This
viewpoint illustrates the relationship between the earlier Bayesian
solutions for GFA \cite{Virtanen12aistats,Gupta12,Damianou12}; they
follow the same general approach presented in
Section~\ref{sec:general}, but our sparsity prior is more advanced.

Perhaps the more interesting interpretation is to consider
\eqref{eq:lrmodel} as the primary model. Then the entries of $\Alp$
are considered as unobserved data describing the groups; $\U$ are the
factor loadings and $\V$ provide the latent factors for the
groups. The mapping from $\Alp$ to the observations, parameterized by
$\Z$ and $\W$, is then merely a set of nuisance parameters. From this
perspective, the earlier models presented for the GFA problem
are very simple. They do not assume any structure between the groups,
but instead draw the association strengths independently. Their results
will naturally still reveal such associations, but not as well as
the proposed model that models them explicitly.

As we later empirically demonstrate, the rank $R$ of the group
association level can typically be very low even for very high-dimensional
data collections with a large number of groups.  This makes it possible
to visually study the associations between the groups, for example via
a scatter plot of the columns of $\U$ for $R=2$. 
We discuss approaches for determining the value of $R$ for practical 
applications in Section 5.4.

\subsection{Inference}

For inference, we use mean-field variational approximation.
We approximate the posterior with a factorized distribution
\[
q(\Theta) = q(\Z)q(\W)q(\Ta)q(\U)q(\V),
\]
where $\Theta=\{ \Z,\W,\Ta,\U,\V \}$, and find the approximation that minimizes the Kullback-Leibler
divergence from $q(\Theta)$ to $p(\Theta|\Y)$.  Equivalently,
this corresponds to estimating the marginal likelihood
$p(\Y)$ with maximal lower bound,
\[
\log p(\Y) \ge L(\Theta)= \int q(\Theta) \log \frac{p(\Theta,\Y)}{q(\Theta)}.
\]
The mean-field algorithm proceeds by updating each of the terms in
turn, always finding the parameters that maximize the expected lower
bound $L(\Theta)$, given all the other factorized distributions.
Since the model uses conjugate priors for everything except $\U$ and
$\V$, the updates for most parameters are straightforward and match
those of other FA models, for example \cite{Klami13jmlr}.  The terms
$q(\U)$ and $q(\V)$ are more complex and hence we derive the updates
for that part in detail below. The VB inference 
is summarized in
Algorithm \ref{algorithm}, and the
parameters of the variational distributions
can be found in the Appendix. An open-source implementation of the model in
R is available as part of the CCAGFA package in CRAN (\url{http://cran.r-project.org/package=CCAGFA}).

\begin{algorithm}[!t]
\begin{algorithmic}
\caption{VB inference of GFA}\label{algorithm}
\STATE Input: initialized $q(\W),q(\Z),q(\Ta)$, and either $q(\Alp)$ or $\U$ and $\V$.
 \WHILE{not converged}
  \STATE{Check for empty factors to be removed}
  \STATE{$q(\W)  \gets \prod_{m=1}^M \prod_{j=1}^{D_m} \mathcal{N}(\w^{(m)}_{:,j}|\m^{(w)}_{m,j},\Sig^{(w)}_m)$}
  \STATE{$q(\Z)  \gets \prod_{i=1}^N \mathcal{N}(\z_{i}|\m^{(z)}_i,\Sig^{(z)})$}
  \IF{full-rank GFA ($R=\min(M,K)$)}
  	\STATE{$q(\Alp) \gets \prod_{m=1}^M\prod_{k=1}^K \G(a_{m,k}^\alpha, b_{m,k}^\alpha)$}
  \ELSE
    \STATE{$\U,\V \gets \arg \max_{\U,\V} L(\Theta)_{\U,\V}$}
  	\STATE{$\langle \Alp \rangle \gets \exp(\U\V^\top)$}
  \ENDIF
  \STATE{$q(\Ta) \gets \prod_{m=1}^M \mathcal{G}(\tau_m| a_m^{\tau}, b_m^{\tau})$}
 \ENDWHILE
\end{algorithmic}
\end{algorithm}

For $q(\U)$ and $q(\V)$ we use fixed-form distributions, that is, we
choose point distributions $q(\U)=\delta_{\U}$ and
$q(\V)=\delta_{\V}$, and optimize the lower bound
numerically\footnote{To keep the notation clean we assume the $\muu$
  and $\muv$ have been appended as parts of $\U$ and $\V$,
  respectively.}. The bound as a function of $\U$ and $\V$ is given by
\begin{align}
L(\Theta)_{\U,\V} =& \sum_{m,k} \left( D_m \uu_{m}^\top \vv_{k} - \langle \W^{(m)}\W^{(m)^\top} \rangle_{k,k} e^{\uu_{m}^\top \vv_{k}} \right) \nonumber\\
&+ 2\log p(\U,\V) , \label{eq:logalpha}
\end{align}
where $\langle \W^{(m)}\W^{(m)^\top} \rangle$ denotes the second
moment of $\W^{(m)}$ with respect to $q(\W^{(m)})$, and $\log p(\U,\V)$
collects the prior terms affecting the factorization.  Since the parameters $\U$ and
$\V$ are highly coupled, we optimize \eqref{eq:logalpha} jointly with
second order approximate gradient method (L-BFGS), using
the gradients
\begin{align*}
\frac{\delta L}{\delta \U} & = \A\V +\frac{\delta \log p(\U,\V)}{\delta \U}, & 
\frac{\delta L}{\delta \V} & = \A^\top\U + \frac{\delta \log p(\U,\V)}{\delta \V},
\end{align*}
where $\A = \mathbf{D}\mathbf{1}^\top - \exp(\U\V^\top)$. 
Full variational inference over these parameters would also be possible
\cite{Dikmen12}, but we did not consider it necessary for the
practical applications.

An interesting special case is obtained when $R=\min(M,K)$. 
Then the factorization is not low-rank, but instead we can find
a unique optimal solution for \eqref{eq:logalpha} as
\[
\alpha_{m,k} = \frac{D_m}{\langle \W^{(m)}\W^{(m)^\top} \rangle_{k,k}},
\]
assuming $\lambda$ is negligibly small.
This is identical to the variational update for a model that draws
$\alpha_{m,k}$ from a gamma prior with the parameters approaching zero
(the uniform distribution limit of gamma). This is the prior
used by some of the earlier GFA solutions \cite{Virtanen12aistats,Damianou12}, and hence we get the earlier models as special cases of
ours.

The inference scales linearly in $M$, $D$ and $N$, and has cubic complexity
with respect to $K$. In practice, it is easily applicable for large
data sets as long as $K$ is reasonable (at most hundreds). However, during inference 
empty factors may occur and in this case they can be removed from the model to speed
up and stabilize computation\footnote{We remove the $k$th factor from the model if 
$c_k = \sum_{i=1}^N \langle \z_{i,k} \rangle ^2/N < 10^{-7}$.}.

\subsection{Predictive inference}

Even though the GFA model is in this work formulated primarily as a
tool for exploring relationships between variable groups, it can
readily be used also as a predictive model. In this prediction setting 
new (test) samples are observed for a subset of groups and the task is 
to predict unobserved groups based on observed data.

For simplicity of presentation, we assume that only the $m$th group is unobserved.
Then the missing data are represented by the
predictive distribution $p(\X^{(m)^*} | \Y^{-(m)^*} )$,
where $\Y^{-(m)^*}$ denotes partially observed test data consisting of all the other
groups. However, this distribution involves marginalization over both $\W$ and $\Z$ that is 
analytically intractable and hence we need to approximate it. In particular, given $\Y^{-(m)^*}$, $q(\W)$ and 
$q(\Ta)$, we learn the approximate posterior distribution for the latent variables $q(\Z^*)$ corresponding to $\Y^{-(m)^*}$ and approximate the mean of the predictive distribution as
\begin{align}
\langle \X^{(m)^*} | \Y^{-(m)^*} \rangle & = \langle \Z^*\W^{(m)} \rangle_{q(\W^{(m)}),q(\Z^*)} \notag \\
& = \Y^{-(m)^*} \boldsymbol{\mathrm{T}} \langle \W^{-(m)\top}\rangle \Sigmab^{-1} \langle\W^{(m)}\rangle, \label{eq:pred}
\end{align}
where $\boldsymbol{\mathrm{T}}=\text{diag}(\{ \langle \tau_j \rangle \I_{D_j}\}_{j\neq m})$ and $\Sigmab = \I_K + \sum_{j\neq m} \langle \tau_j \rangle \langle \W^{(j)}\W^{(j)^\top} \rangle$. In the experiments we use 
this mean value for prediction. 

\section{Related work}
\label{sec:related}

The GFA problem and our solution for it are closely related to several
matrix factorization and factor analysis techniques. In the following,
the related work is discussed from two perspectives. First we cover
other techniques for solving the group factor analysis problem or
its special cases. Then we proceed to relate the proposed solution to
multiple regression, which is a specific use-case for GFA.

\subsection{Factor models for multiple groups}

For a single group, $M=1$, the model is equivalent to Bayesian
principal component analysis \cite{Bishop99nips,Ilin10}; all of the factors are active for
the one group and they describe the variation with linear components.  We
can also implement sparse FA with the model, by setting $M=D$ so that
each group has only one variable. The residual noise has independent
variance for each variable, and the projections become sparse because of
the ARD prior.

For two groups, $M=2$, the model is equivalent to Bayesian CCA and
inter-battery factor analysis \cite{Klami13jmlr}; some factors model
the correlations whereas some describe the residual noise within
either group.

Most multi-set extensions of CCA, however, are not equivalent
to our model. For example, Archambeau et al. \cite{Archambeau08}
and Deleus et al. \cite{Deleus11} extend CCA for $M>2$, but instead of GFA
they solve the more limited problem of multiple-battery
factor analysis \cite{McDonald70,Browne80}. The MBFA models
provide one set of factors that describe the relationships
between \emph{all} groups, and then model the variation specific
to each group either with a free covariance matrix or a separate
set of factors for that group. Besides the multi-set extensions
of CCA, also the probabilistic interpretation of
sparse matrix factorization \cite{Qu11}, and
the JIVE model for integrated analysis of multiple
data types \cite{Lock11,Ray13} belong to the
family of MBFA models. These models differ in their priors, parameterization
and inference, but are all conceptually equivalent.

In recent years, a number of authors have independently proposed
solutions for the GFA problem. They all follow the general solution
outline presented in Section~\ref{sec:general} with varying techniques
for obtaining the group-wise sparsity.
Common to all of them is that they do
not explicitly model the relationships between the groups, but instead
assume that the choice of whether a factor describes variation in
one particular group can be made independently for all factor-group
pairs. This holds for the non-Bayesian solutions of multi-view sparse
matrix factorizations \cite{Jia10} and the group lasso penalty
variant of SCA \cite{VanDeun11}, as well as for the earlier Bayesian
GFA models \cite{Virtanen12aistats,Gupta12,Damianou12}. Compared to these, our model explicitly describes the
relationships between the groups, which helps especially for setups
with a large number of groups. Finally, we get the sparsity priors
of \cite{Virtanen12aistats} and \cite{Damianou12} as special cases of our model.

\subsection{Factor regression and group-sparse regression}

The GFA problem can also be related to supervised learning, by considering
one of the groups as dependent variables and the others as explanatory
variables. For just one group of dependent variables (that is, $M=2$ in total),
GFA is most closely related
to a setting called factor regression \cite{West03}.
It shares the goal of learning a set of latent factors that
are useful for predicting one group from the other. For
more recent advances of factor regression models, see \cite{Chen10,Rai09}.
By splitting the explanatory variables into multiple groups,
GFA provides a group-wise sparse alternative for these models.
Assuming the split corresponds to meaningful prior information
on the structure of the explanatory variables, this will usually 
(as demonstrated in experiments)
reduce overfitting by allowing the model to ignore group-specific
variation in predictions.

Other models using group-wise sparsity for regression have also been
presented, most notably group lasso \cite{Yuan06,raey13} that uses a
group norm for regularizing linear regression. Compared to GFA, lasso
lacks the advantages of factor regression; for multiple output cases
it predicts each variable independently, instead of learning a latent
representation that captures the relationships between the inputs and
outputs. GFA has the further advantage that it learns the predictive
models for not only all variables but in fact for all groups at the
same time. Given a GFA solution one can make predictions for arbitrary
subsets of the groups given another subset, instead of needing to
specify in advance the split into explanatory and dependent variables.

\section{Technical demonstration}
\label{sec:toy}

In this section we demonstrate the proposed GFA model on artificial
data. To illustrate the strengths of the proposed method we compare it
with Bayesian implementations of the most closely related factor models,
always using a variational approximation for inference and ARD for
complexity control also for the competing methods.
In particular, we
compare against the regular factor analysis (FA) and its sparse version (sFA) to
show that one should not completely ignore the group information. We
also compare against
MBFA, to demonstrate
the importance of modeling also relationships between subsets of the
groups, and against the GFA solution of \cite{Virtanen12aistats},
obtained as a special case of the proposed model by setting $R=\min(M,K)$, as an example of a
method that makes the group-factor activity decisions independently
for each pair. For MBFA we use two different implementations depending
on the setup; for low-dimensional data we model the
group-specific variation with full-rank covariance, whereas for
high-dimensional data we use a separate set of factors for each group;
see Klami et al. \cite{Klami13jmlr} for discussion on these two alternatives
for the special case of $M=2$.

We also compare against SCA \cite{VanDeun11}, a non-Bayesian solution
for the GFA problem using group lasso penalty, using the same initial
$K$ as for the Bayesian models, with 5-fold cross-validation for the
the group lasso regularization parameter. For predictive inference we
compute point estimates for the latent variables of the test samples.

\subsection{Evaluation}

For evaluating the quality of the models we use an indirect measure of
predictive accuracy for left-out groups, based on the intuition that
if a factor analysis model is able to make accurate predictions it
must have learned the correct structure. In particular, given
$M$ groups we will always compute the test data predictive error for
each of the groups one at a time, using the rest of the groups as
explanatory variables.

Since we use a regression task for measuring the quality, we will
also compare GFA against alternative standard solutions for multiple output
regression, in addition to the alternative factor models mentioned
in the previous section. In particular, we will show comparison
results with group lasso \cite{raey13} and simple regularized multiple output
linear regression (MLR) model that ignores the group structure. For MLR the prediction is obtained as $\X^{(m)^*} = \Y^{-(m)^*} \B$, where the weight 
matrix is given by $\B= (\Y^{-(m)^\top} \Y^{-(m)} + \gamma \I)^{-1}\Y^{-(m)^\top} \X^{(m)}$, and $\gamma$ is a regularization parameter.
For this model we learn a separate model for each choice of the
dependent variable groups, which results in $M$-fold increase in
computational cost compared to GFA that learns all tasks
simultaneously. Furthermore, we validate for the regularization parameters
via 10-fold cross-validation within the training set, which further
increases the computational load.

\subsection{Beyond factor analysis and MBFA}

\begin{figure*}[ht!]
 \centering
 \includegraphics[width=0.95\textwidth]{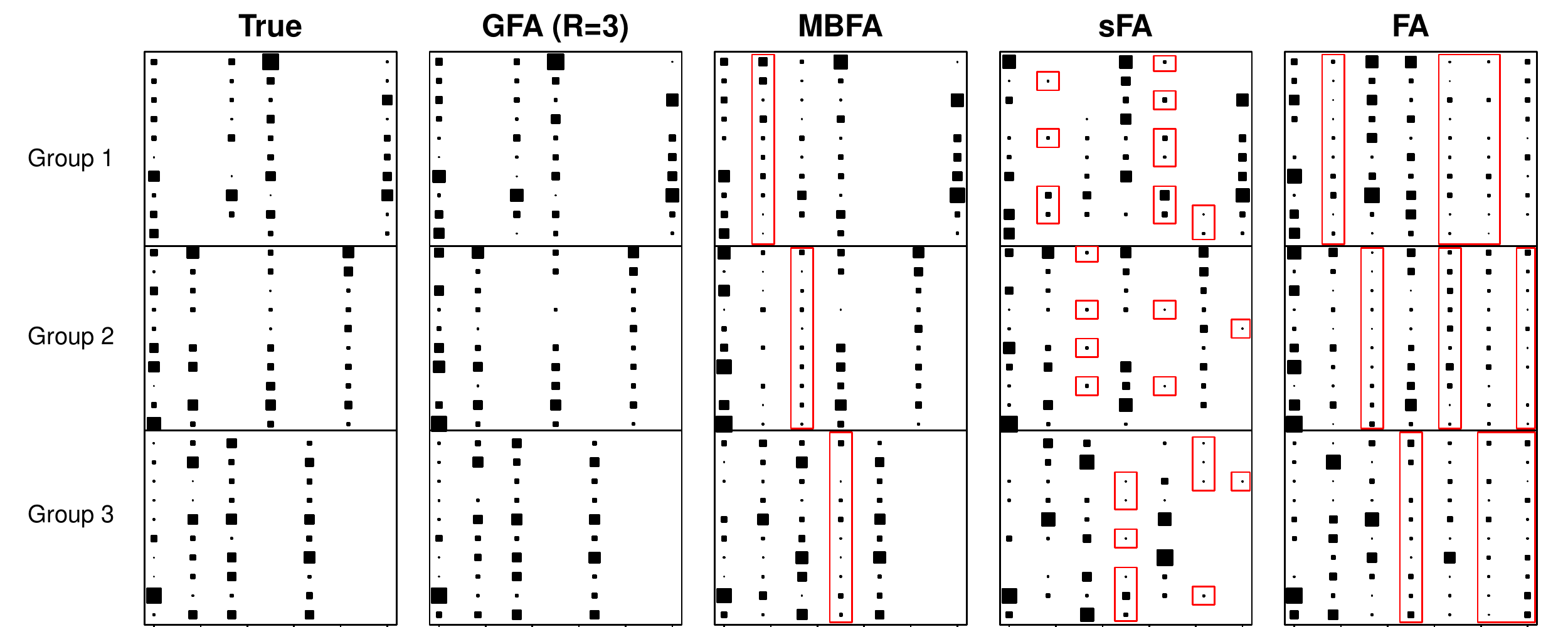} 
\caption{The projections $\W$ for the true generating model,
the proposed GFA solution (using full rank, $R=3$), and comparison methods 
multiple-battery factor analysis (MBFA), sparse factor analysis (sFA),
and regular factor analysis (FA).
GFA finds the true solution, whereas the other methods report
spurious correlations (red boxes). The three sub-blocks in each matrix correspond
to the three groups of variables, and the areas of the black patches
indicate the absolute values of the corresponding weights.
}
\label{fig:W}
\end{figure*}

We generated $N=100$ samples from a GFA model with $D=30$ split into
$M=3$ equally-sized groups. The true generative model had $K=7$
factors, including one factor specific for each group as well as for
each possible subset between the groups.  Figure~\ref{fig:W} shows the
true model as well as the solutions found by the proposed GFA model
(using $R=\min (M,K) = 3$), MBFA, and both regular and sparse FA. The
proposed model finds the correct structure, whereas MBFA suggests
spurious correlations; each factor describing correlation between just
two of the groups is falsely indicating activity also in the third
one, according to the MBFA specification. The regular FA result is
simply considerably noisier overall, while sparse FA suffers from a
few false factor loadings. For this simple demonstration SCA learns
the same result as GFA, after manually optimizing the thresholding of component
activity.  For all methods we set $K$ to a sufficiently large number,
allowing ARD to prune out unnecessary components, chose the best
solution by comparing the lower bounds of $10$ runs with random
initializations, and for illustrative purposes ordered the components
based on their similarity (cosine) with the true ones.

The conclusion of this experiment is that the proposed method indeed
solves the GFA problem, whereas the MBFA and
FA solutions do not provide as intuitive and interpretable
factors. For this data GFA with $R<3$ (not shown) fails to unveil the 
underlying (full-rank) structure. Instead, the loadings lie between those of GFA and FA of Figure~\ref{fig:W},
which is understandable since FA is closely related to GFA with $R=0$.

\subsection{Performance for several groups}

\begin{figure*}[ht!]
 \centering
 \includegraphics[width=1.0\textwidth]{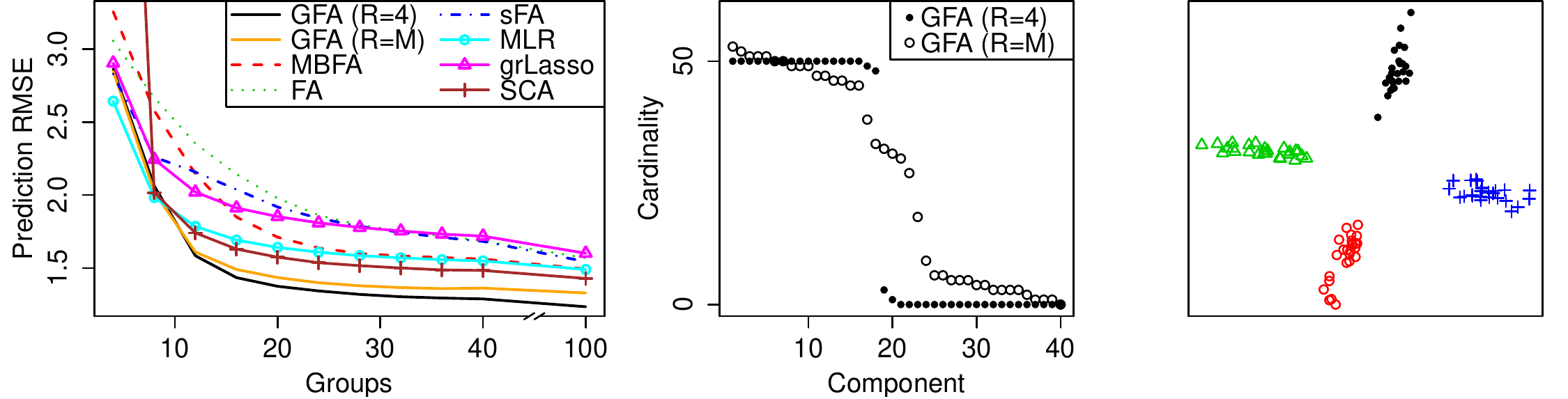}
\caption{\textbf{Left}: GFA is considerably better in modeling the relationships between
  the groups compared to MBFA, FA and SCA; the y-axis shows the average
  prediction error for a new group on the test data, providing an
  indirect quality measure. For large $M$ explicitly modeling the
  relationships between the groups ($R$=4) helps compared to earlier
  solutions that make the activity decisions independently ($R=\min(M,K)$,
  here $R$=$M$).
  GFA also outperforms standard regression models in the prediction task,
  with the exception of MLR for very low dimensional data.
  \textbf{Middle}: For $M=100$ the $R=4$ solution correctly finds the $K=18$
  factors that are all active in $50$ groups, whereas the earlier
  solution splits some of the factors into several ones.
  \textbf{Right}: Scatter-plot of matrix $\U$ for
  a $R$=2 solution, illustrating clearly the underlying structure of
  the data. Here the symbols and colors indicate the ground truth types of
the groups that were not available for the algorithm during learning.}
\label{fig:M}
\end{figure*}

Next we studied how well the proposed solution scales up for more
groups.
We generated $N=30$ samples from a GFA model with $K=18$ true
factors. We used $D_m=7$ variables
for each group and let the number of groups grow from $M=4$ to $M=100$.
In other words, the total dimensionality of the data grew from $D=28$ to
$D=700$.
The variable groups were divided into four types of equal size, so that
the groups within one type had the same association strengths for
all factors. This implies a low-rank structure for the associations.

\begin{figure*}[t!]
 \centering
 \includegraphics[width=0.95\textwidth]{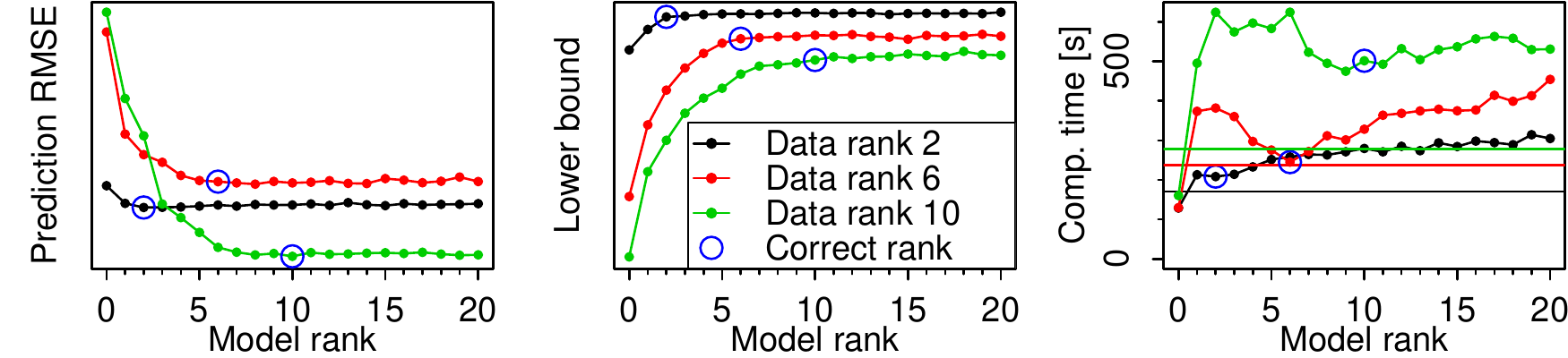}
\caption{The 5-fold cross-validation prediction performances ({\bf left}), 
lower bounds ({\bf middle}) and computation times ({\bf right}) for three
different artificial data sets as a function of the model rank $R$. 
Both approaches provide a reliable way for choosing the rank $R$. 
In the illustrated example the chosen model ranks using cross-validation are 
$2$, $8$, and $10$ (correct: $2$, $6$ and $10$).
By monitoring the lower bounds the true ranks, used to generate the data sets,
can be approximately detected from the figure.
For clarity, the vertical positions of prediction and lower bound curves were
here altered, retaining the relative changes.
The {\bf right} subplot illustrates the computational cost as a function
of the rank $R$, compared against the full-rank solution shown as horizontal
lines.}
\label{fig:lb}
\end{figure*}

For validation we used average leave-one-group-out prediction error,
further averaged over 50 independent data sets.
Figure~\ref{fig:M} (left) shows that the proposed model outperforms
the other FA methods by a wide margin.
For small $M$ the difference between
the $R=4$ and $R=\min(M,K)$ solutions are negligible, since a small
number of factor-group association strengths can just as well be
selected independently. For large $M$, however, explicitly modeling
the relationships pays off and results in better predictions. The prediction
errors for GFA models with rank $R$ between 2 and 10 are very similar, and
hence for clarity, only one ($R$=4) is shown in Figure~\ref{fig:M}.
GFA also outperforms SCA, an alternative group-sparse factor model,
for all cases except $M=8$, for which the two methods are equal.
For cases with only 4 or 8 groups multiple linear regression is the most accurate
method, but for all other cases, when the total dimensionality increases, GFA clearly outperforms also these supervised
methods.

Besides showing better prediction accuracy, the low-rank solution (here $R=4$)
captures the underlying group structure better than the naive model of
$R=\min(M,K)$. Figure~\ref{fig:M} (middle) compares the two by plotting the
number of groups active in each factor for the case with $M=100$. There are
$K=18$ factors that should each be active in $50$ groups.  With $R=4$
we find almost exactly the correct solutions, whereas the naive model
finds $K=40$ factors, many of which it believes to be shared by
only $5-40$ groups. In other words, it has split some real factors
into multiple ones, finding a local optimum, due to making all the
activity decisions independently. For illustrative purposes, the components
were considered active if the corresponding $\Alp$-values were below 10.
The cardinality plot is sensitive to the threshold, but the
key result stays the same regardless of the threshold: inferring the
component activities independently results in a less accurate model.

Finally, Figure~\ref{fig:M} (right)
illustrates the relationships between the groups as a scatter plot of
the latent factors for the $R=2$ solution, revealing clearly the
four types of variable groups.

\subsection{Choosing the model rank}

GFA contains a free parameter $R$ and this value needs to be specified by the user.
Acknowledging that choosing the correct model order is in general a difficult
problem, we resort to demonstrating two practical approaches that seem to work
well for the proposed model. The first choice is $L$-fold cross-validation within the training
set, using the predictive performance for left-out groups as the validation criterion.

A computationally more efficient alternative is to use the 'elbow'-principle for
the variational lower bounds $L(\Theta)$ computed for different choices of $R$.
The bounds improve monotonically\footnote{Given that the constant terms in the
priors of $\U$ and $\V$ are ignored.}
as a function of $R$,
but for typical data sets the growth rate rapidly
diminishes after the correct rank, producing an 'elbow'.

\begin{figure*}[t!]
 \centering
\includegraphics[width=1.00\textwidth]{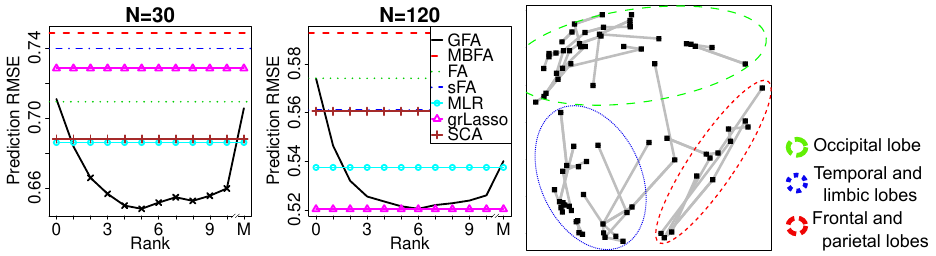}
\caption{{\bf Left and middle:} Predictive errors (smaller is better) for $N=30$ and $N=120$ training samples. For $N=30$ the proposed model outperforms the competitors for almost all ranks (the crosses indicate statistically significant difference compared to all competitors), with $R=5$ providing the best results. For $N=120$, GFA for ranks $3-10$ is significantly better than alternative factor models and MLR, equaling the predictive performance of group lasso for correctly chosen rank $R$. {\bf Right:} Illustration of the group factor loadings for $R=2$. The model automatically discovers three groups of ROIs that can be identified as the occipital lobe, temporal and limbic lobes, and frontal and parietal lobes.}
\label{fig:brainPred}
\end{figure*}

We tested both of these principles for 50 independent artificial data sets generated from
the GFA model with parameters $N=50$, $K=30$, $M=50$ and $D_m=10$, for three
different data ranks: $R=\{2,6,10\}$, representing the kinds of values one would
typically expect for real applications. The prediction and lower bound curves as
a function of model rank are shown for a representative example in Figure~\ref{fig:lb}.
In the 5-fold cross-validation the correct rank was found correctly with over half of
the data sets, and the most common error was overestimating the rank by one.
The computationally lighter 'elbow'-principle allows the analyst to choose roughly
the correct rank by simply comparing the lower bounds.

The rank $R$ influences also the computation time, as illustrated
in Figure~\ref{fig:lb}.
The computation time increases roughly linearly as a function
of $R$, but for ranks smaller than the optimal the algorithm requires
more iterations for convergence which slows it down.
In this experiment, the low-rank model is slower than the special 
case updating $\Alp$ in closed form, but only by a small factor.
In the real data experiments reported in Sections~\ref{sec:brain} and \ref{sec:bio} the low-rank variant was slightly faster to compute
for all the tested values of $R$; the relative time spent on updating
$\Alp$ becomes small for larger data, and the low-rank models prune
out excess factors faster.

In the remaining experiments we do not explicitly select a particular value for $R$,
but instead present the results for a range of different values. This is done to
illustrate the relative insensitivity of the model for the precise choice; for both
real-world applications a wide range of values outperform the alternative models
and also the special case of $R=\min(M,K)$, implying that picking exactly the right rank is not crucial.

\section{Analysis of brain regions}
\label{sec:brain}

Functional MRI experiments are commonly used to study the interactions
(or connectivity) between brain regions of interest (ROIs)
\cite{Friston94,Smith09}.
One way to learn these interactions is based on calculating correlations between
individual fMRI BOLD (blood-oxygen-level dependent) signals using PCA or FA models \cite{Friston93}.

We applied GFA to brain imaging data to analyze
connections between multiple brain ROIs, using
fMRI data recorded during natural stimulation with a piece of music
\cite{Alluri12}, for $N=240$ time points covering two seconds
each. We have $11$ subjects, and we computed a separate GFA model for
each, providing $11$ independent experiments.
We chose $M=81$ ROIs, some of which are lateralized, using the FSL software 
package \cite{Jenkinson12}, based on
the Harvard-Oxford cortical and subcortical structural atlases \cite{Desikan06} and the probabilistic cerebellar atlas \cite{Diedrichsen09}. 
Further, we divided the voxels in these regions to
$\sum_m D_m = 676$ local uniformly distributed supervoxels by spatial averaging. In the end,
each ROI was represented on average by eight such supervoxels and each supervoxel contained on average 120 voxels. These ROIs and
their corresponding dimensionalities are given in Supplementary material
available at \url{http://research.ics.aalto.fi/mi/papers/GFAsupplementary.pdf}.

For quantitative comparison we again use 
leave-one-group-out prediction, predicting the activity of each ROI
based on the others for unseen test data.
We set aside half of the data as
test set and train the models varying the amount of training data for
$K=100$. The prediction errors are given in Figure
\ref{fig:brainPred}, averaged over the ROIs and $11$ subjects. The
proposed solution outperforms (Wilcoxon signed-rank test, $p<10^{-6}$)
all other factor models for a wide range of ranks, from $R=2$ to $R=10$, and
in particular is also clearly better than the special case with $R=\min(M,K)$
\cite{Virtanen12aistats}. For these ranks, with $N=30$ training samples, 
GFA also outperforms all the supervised regression models, whereas for
a larger training set, $N=120$, group lasso provides comparable
accuracy.

Figure~\ref{fig:brainPred} shows also a visualization of $\U$ for $R=2$, averaged over all subjects using all observations. 
Since $\U$ is not identifiable, before averaging we projected each $\U$ to a common coordinate system. Each dot is one ROI and the lines connect each ROI to its spatially closest neighbor (minimum Euclidean distance of supervoxels between the corresponding ROIs)
 to reveal that the model has learned interesting structure despite not knowing anything about the anatomy of the brain. Further inspection reveals that the
model partitions visual areas, frontal areas, and auditory areas as separate clusters.
Note that these results are a demonstration of the model's capability
of discovering structure between groups; for a serious attempt
to discover functional or anatomic
connectivity further tuning that takes into account the properties
of fMRI and the anatomy of the brain should be done.

\section{Analysis of drug responses}
\label{sec:bio}

\begin{figure*}[t!]
 \centering
 \includegraphics[width=1.00\textwidth]{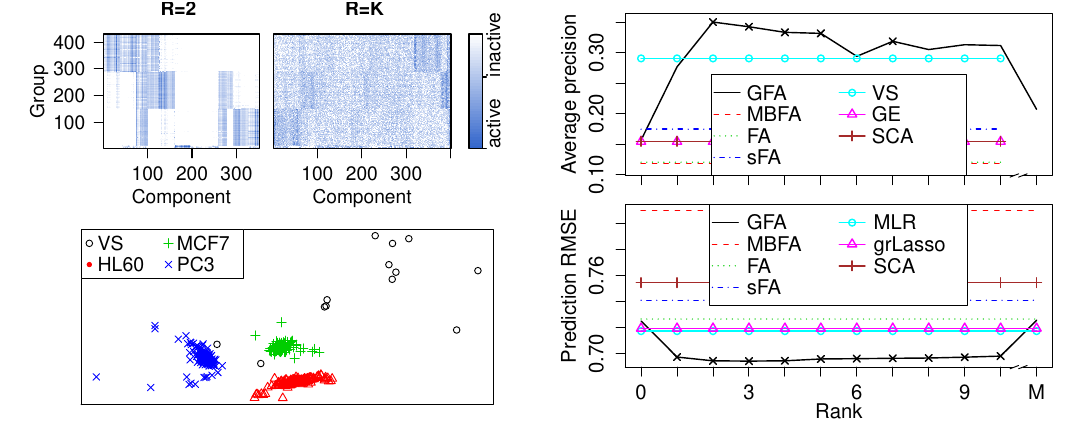}
\caption{{\bf Top left:} The group-factor associations of GFA for $R=2$ and $R=K$ illustrate how the low-rank model reveals much clearer structure. The first 13 groups are the chemical descriptors, followed by 139 feature groups (pathways) for each of the 3 cell lines.  {\bf Top right:} Drug retrieval accuracies (higher is better, the crosses indicate statistical significance); the GFA solution is clearly the best for a wide range of ranks, and it is the only method outperforming retrieval based on the chemical descriptors. Here GE denotes raw gene expression of all three cell lines and VS corresponds to the chemical descriptors.
Note that the accuracy of SCA is almost identical to GE, which makes the two lines hard to distinguish.
{\bf Bottom left:} The group factor loadings $\U$ for $R=2$, showing how the model is able to separate the three cell lines and the chemical descriptors almost perfectly. {\bf Bottom right:} Predictive errors (smaller is better) for the drug response data. GFA with
ranks $R=1,...,10$ outperforms the comparison methods significantly. }
\label{fig:drug}
\end{figure*}

Both chemical descriptors and biological responses can be used to
analyze the properties of drugs. However, the shared variation
between these two views may reveal more of the drug properties than
either one of the views independently \cite{Khan12}.
We applied GFA to the drug response data of \cite{Lamb06},
consisting of the responses of $N=684$ drugs when applied to three
cancer cell lines (\textit{HL60, MCF7} and \textit{PC3}). For the analysis,
we selected the genes found in CP: Biocarta gene
sets\footnote{http://www.broadinstitute.org/gsea/msigdb/collections.jsp},
where the genes are grouped according to functional pathway information.
We preprocessed the data as Khan et al. \cite{Khan12} and left duplicate
genes only in the largest groups, removing groups with less than two genes
left. As in \cite{Khan12}, we augmented the data by adding 76 chemical descriptors
(\textit{VolSurf}) of the drugs, here as separate variable groups. Hence the whole
data contain $M=430$ groups: 13 contain chemical descriptors of the drugs,
whereas 139 groups describe the response in each of the three cell lines.
The feature groups and their dimensionalities are listed in the
Supplementary material available at \url{http://research.ics.aalto.fi/mi/papers/GFAsupplementary.pdf}.
The total data dimensionality is $D=3172$.

Figure~\ref{fig:drug} illustrates GFA results with $K=400$ on this data
(a component amount high enough for GFA with $R<10$). Already
with $R=2$ the model partitions the groups almost perfectly into four
different clusters (bottom left subfigure), one for each of the three cell lines and one for
the chemical descriptors. That is, the model effectively learns the
underlying structure it has no information about.  It also reveals
clearer factor-group association structure (top left subfigure) compared to the earlier
solution with $R=M$ \cite{Virtanen12aistats}. With $R=3$ the four different
clusters can be perfectly separated.

Next we quantitatively validated the performance of the different
solutions, using a drug retrieval task \cite{Khan12}. Using one of the
drugs as a query, we ranked the remaining drugs based on their
similarity with the query, and used an external database of drug
functions for assessing the retrieval results. By comparing the
similarity in the latent space of the models, we can indirectly
evaluate the quality of the representations the models have
learned. It has been shown that the chemical VolSurf descriptors
capture drug functions significantly better than raw gene expression
data for this data \cite{Khan12}, and hence we computed the similarity
measures of all the models based on factors that are active in at
least one chemical descriptor group. For this purpose we thresholded
the activities by regarding components with $\Alp < 100$ as active;
the results were not very sensitive for the exact threshold level.
The retrieval can be quantified by measuring the average precision
\cite{baeza1999modern}, further averaged over all the drugs (separate
retrieval tasks), which is a measure that gives most weight to the
drugs that are retrieved first.  Figure~\ref{fig:drug} (top right)
shows that the proposed solution again outperforms all of the
competing methods for all ranks above zero (Wilcoxon signed-rank test,
$p<10^{-6}$), and for $R=2$ to $R=5$ and $R=7$ it significantly
($p<0.05$) outperforms also the chemical descriptors that are
considerably more accurate than any of the competing
methods. The shown retrieval accuracies are based on the ten most
similar drugs, but the results are consistent for sets of several
sizes.

In addition to the retrieval task, we measured the performance of the models
in a leave-one-group-out prediction task with $N=616$. The average 
errors are shown in Figure~\ref{fig:drug} (bottom right). GFA with $R=1,...,10$ outperforms
all the competing models significantly.

\section{Discussion}

Joint analysis of multiple data sets is one of the trends in machine
learning, and integrated factor analysis of multiple real-valued
matrices is one of the prototypical scenarios for that task.  In
recent years multiple authors have re-discovered the multiple-battery
factor analysis (MBFA) task originating from the early works in
statistics
\cite{McDonald70,Browne80,Tucker58,Browne79}, calling it either
multi-set CCA
\cite{Archambeau08,Deleus11}, or simply as a model for integrated analysis of
multiple data sources \cite{Lock11,Ray13}. Despite varying technical
details, all of these models can be seen as FA models
with two sets of factors: one set describes dependencies between
all of the variable groups, whereas the other set
describes, or explains away, variation specific to each group.

The group factor analysis problem formulated in this article,
extending the preliminary treatment in \cite{Virtanen12aistats},
differs from the MBFA models in one crucial aspect. Instead of
only modeling relationships between \emph{all} of the groups, we also
introduce factors that model relationships between any subset of them. While some other recent works \cite{Jia10,VanDeun11,Gupta12,Damianou12} have
also addressed the same problem, in this paper the GFA setup is for
the first time introduced explicitly, putting it into its statistical
context. We described a general solution principle that covers the
earlier solutions, identifying the structural sparsity prior as the
key element. We then presented a more advanced
sparsity prior that results in a novel GFA solution:
Instead of choosing the activities of each group-factor pair
independently, we explicitly
model the relationships between the groups with another linear
layer. Our model hence directly provides factor loadings also
between the groups themselves, which was exactly the original
motivation for the GFA problem. Our preliminary model
\cite{Virtanen12aistats} is a special case with \emph{a priori}
independent loadings.

We showed, using artificial data, how the GFA problem and solution
differ from the MBFA-problem and classical FA. We also demonstrated
that, especially for a large number of groups or data sets, it pays
off to explicitly model the relationships between the groups.
Finally, we applied the model on two real-world exploratory analysis
scenarios in life sciences. We demonstrated that the model is 
applicable to connectivity analysis of fMRI data, as well as for
revealing structure shared by structural description of drugs and
their response in multiple cell lines. These demonstrations
illustrated the kinds of setups the GFA is applicable for, but should
not be considered as detailed analyses of the specific application problems. 

Besides showing that the proposed model solves the GFA problem
considerably better than the alternatives MBFA, FA and SCA \cite{VanDeun11}, the
empirical experiments revealed that there is a qualitative difference
between the proposed model having the more advanced structural
sparsity prior and the earlier GFA solutions such as \cite{Virtanen12aistats}.
Even though the earlier models also solve the GFA problem reasonably
well, they are outperformed by supervised regression models in predictive
tasks. The proposed solution with a low-rank model for the group association
strengths is clearly more accurate in prediction tasks and, at least
for small training sets, outperforms also dedicated regression models
trained specifically to predict the missing groups.
This is a strong result for a model that does not know in advance
which groups correspond to explanatory variables and which to the dependent
variables, but that instead learns a single model for all possible
choices simultaneously.

The model presented here is limited to scenarios where each training
sample is fully observed. Support for missing observations could be
added using the fully factorized variational approximation used for
PCA and collective matrix factorization with missing data
\cite{Ilin10,Klami14iclr}.  A similar approach can also be used for
semi-paired setups where some samples are available only for some groups
\cite{Chen12}, by filling in the remaining groups by missing
observations. Empirical comparisons on these are left for future work.
Another possible direction for future work concerns more justified
inference for the rank parameter $R$; even though the experiments
here suggest that the method is robust to the choice, the method
would be more easily applicable if it was selected automatically.

\section*{Acknowledgment}
\addcontentsline{toc}{section}{Acknowledgment}

We thank the Academy of Finland (grant numbers 140057, 266969, and
251170; Finnish Centre of Excellence in Computational Inference
Research COIN), the aivoAALTO project of Aalto University, and Digile
ICT SHOK (D2I programme) for funding the research.  We would also like
to thank Suleiman A. Khan for his help with the biological
application, and Enrico Glerean for his help with the neuroscience
application. We acknowledge the computational resources provided by
Aalto Science-IT project.

\appendix

\section{Variational inference}
\label{app:updates}

The latent variables are updated as $q(\Z) = \prod_{i=1}^N \mathcal{N}(\z_{i}|\m^{(z)}_i,\Sig^{(z)}) $, where
\begin{align}
\Sig^{(z)} &= \left( \I_k + \sum_{m=1}^M\langle \tau_m\rangle\langle \W^{(m)}  \W^{(m)\top}\rangle \right)^{-1} \nonumber\\
\m^{(z)}_i &= \sum_{m=1}^M\Sig^{(z)} \langle \W^{(m)}\rangle  \langle \tau_m \rangle \x^{(m)}_{i}  . \nonumber
\end{align}

The projection matrices are updated as
$q(\W) =  \prod_{m=1}^M \prod_{j=1}^{D_m} \mathcal{N}(\w^{(m)}_{:,j}|\m^{(w)}_{m,j},\Sig^{(w)}_m)$,
where
\begin{align}
\Sig^{(w)}_m &= \left(\langle\tau_m\rangle\sum_{i=1}^N\langle \z_{i} \z_{i}^\top \rangle + \langle \overline{\overline{\Alp}}_m \rangle \right)^{-1} \nonumber\\
\m^{(w)}_{m,j} &= \Sig^{(w)}_m\langle\tau_m\rangle\left(\sum_{i=1}^N x^{(m)}_{ij} \langle \z_{i}\rangle\right) , \nonumber
\end{align}
and $\overline{\overline{\Alp}}_m$ is the $m$th row of $\Alp$ transformed into a diagonal $K\times K$ matrix. 

Noise precision $q(\Ta) = \prod_{m=1}^M \mathcal{G}(\tau_m| a_m^{\tau}, b_m^{\tau})$ parameters are updated as
\begin{align*}
 a_m^{\tau} &= a^{\tau} + \frac{D_mN}{2} \\
 b_m^{\tau} &= b^{\tau} + \frac{1}{2}\sum_{i=1}^N \left\langle (\x^{(m)}_{i}- \W^{(m)\top} \z_{i} )^2\right\rangle. 
\end{align*}

Finally, for the low-rank model, $\Alp=e^{\U \V^\top + \muu \mathbf{1}^\top + \mathbf{1}\muv^\top}$ is updated by optimizing the lower bound numerically. The bound as a function of $\U$ and $\V$ is given by
\begin{align}
 &\sum_{m,k} D_m \log(\alpha_{m,k})  - \langle \W^{(m)}\W^{(m)^\top} \rangle_{k,k} \alpha_{m,k} \nonumber\\
& - \lambda (\text{tr}(\U^\top\U) + \text{tr}(\V^\top\V))  .\label{eq:logalpha} \nonumber
\end{align}
The gradients w.r.t. the cost function are given as
\begin{align*}
\frac{\delta L}{\delta \U} & = \mathbf{A}\V + \lambda\U,&
\frac{\delta L}{\delta \muv} & = \mathbf{A}\mathbf{1}, \\
\frac{\delta L}{\delta \V} & = \mathbf{A}^\top\U + \lambda\V, &
\frac{\delta L}{\delta \muv} & = \mathbf{A}^\top\mathbf{1}, 
\end{align*}
where $\mathbf{A} = \mathbf{D}\mathbf{1}^\top - \exp(\U\V^\top + \muu\mathbf{1}^\top + \mathbf{1}\muv^\top)$.

With full rank \cite{Virtanen12aistats} the ARD parameters are updated as $q(\Alp)=\G(a^\alpha_{m},b^\alpha_{mk})$, where
\begin{align*}
a^\alpha_{m} &= a^\alpha + \frac{D_m}{2} \\
b^\alpha_{mk} &= b^\alpha + \frac{\w^{(m)\top}_k\w^{(m)}_k}{2} .
\end{align*}

\end{document}